\documentclass[10pt,twocolumn,letterpaper]{article}

\usepackage[table,xcdraw,dvipsnames]{xcolor}
\usepackage{makecell}
\usepackage[algo2e]{algorithm2e}
\usepackage{algorithm}
\usepackage{listings}
\usepackage{multirow}
\usepackage{soul}
\usepackage{wrapfig}
\usepackage{amsmath}
\usepackage{bm}
\usepackage{tabularx}

\usepackage[pagenumbers]{cvpr}      

\def\ie{\emph{i.e.}}
\def\eg{\emph{e.g.}}

\definecolor{cvprblue}{rgb}{0.21,0.49,0.74}
\usepackage[pagebackref,breaklinks,colorlinks,allcolors=cvprblue]{hyperref}

\def\paperID{***}  
\def\confName{CVPR}
\def\confYear{2025}

\title{ResCLIP: Residual Attention for Training-free Dense Vision-language Inference}

\author{
Yuhang Yang$^{*}$\space\space\space\space Jinhong Deng$^{*}$\space\space\space\space Wen Li$^{\dagger}$ \space\space\space\space Lixin Duan \\
    University of Electronic Science and Technology of China\\
{\tt\small \{yvhyang, jhdengvision, liwenbnu, lxduan\}@gmail.com}
}

\newcommand{\tablestyle}[2]{
    \setlength{\tabcolsep}{#1}
    \renewcommand{\arraystretch}{#2}
}

\begin{document}
\maketitle

\vspace{-12pt}
{\renewcommand{\thefootnote}{}
\footnotetext{\hspace{-6pt}$^{*}$Yuhang Yang and Jinhong Deng contributed equally.

$^{\dagger}$Corresponding author.
}}

\begin{abstract}

While vision-language models like CLIP have shown remarkable success in open-vocabulary tasks, their application is currently confined to image-level tasks, and they still struggle with dense predictions. Recent works often attribute such deficiency in dense predictions to the self-attention layers in the final block, and have achieved commendable results by modifying the original query-key attention to self-correlation attention, (e.g., query-query and key-key attention). However, these methods overlook the cross-correlation attention (query-key) properties, which capture the rich spatial correspondence. In this paper, we reveal that the cross-correlation of self-attention in non-final layers of CLIP also exhibits localization properties. Therefore, we propose the Residual Cross-correlation Self-attention (RCS) module, which leverages the cross-correlation self-attention from intermediate layers to remold the attention in the final block. The RCS module effectively reorganizes spatial information, unleashing the localization potential within CLIP for dense vision-language inference. Furthermore, to enhance the focus on regions of the same categories and local consistency, we propose the Semantic Feedback Refinement (SFR) module, which utilizes semantic segmentation maps to further adjust the attention scores. By integrating these two strategies, our method, termed \textbf{ResCLIP}, can be easily incorporated into existing approaches as a plug-and-play module, significantly boosting their performance in dense vision-language inference. Extensive experiments across multiple standard benchmarks demonstrate that our method surpasses state-of-the-art training-free methods, validating the effectiveness of the proposed approach.
Code is available at \url{https://github.com/yvhangyang/ResCLIP}.

\end{abstract}
\section{Introduction}
\label{sec:intro}
\begin{figure}
    \centering
    \tablestyle{0pt}{0.1}
    \includegraphics[width=\linewidth]{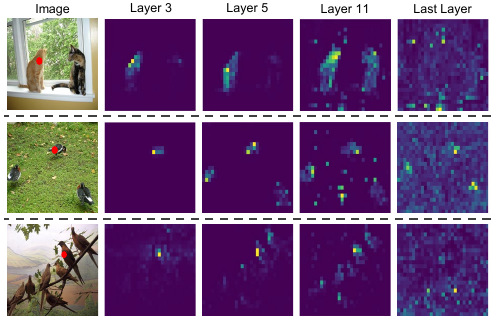}
    \vspace{-6mm}
    \caption{ The attention visualization from different layers of CLIP~\cite{radford2021learning} model. The images are sampled from PASCAL VOC~\cite{everingham2015pascal} dataset. 
    }
    \label{fig:differ_layer_attention}
    \vspace{-6mm}
\end{figure}

\begin{figure*}
    \centering
    \tablestyle{0pt}{0.1}
    \includegraphics[width=0.85\textwidth]{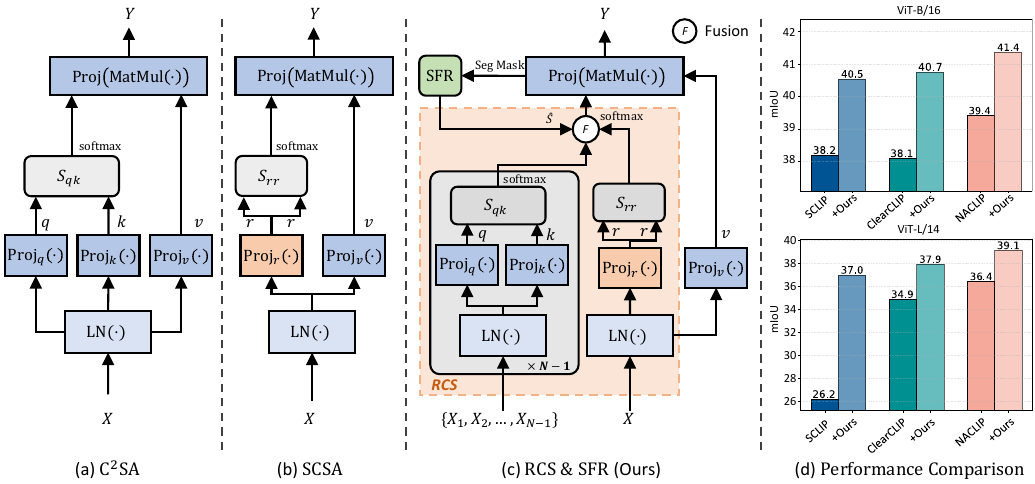}
    \vspace{-4mm}
    \caption{(a) Cross-correlation self-attention ($\text{C}^{2}$SA). The query and key are mapped by different project matrices. The attention is obtained by matrix multiplication between query and key. (b) Self-correlation self-attention (SCSA). The attention is calculated by the self-correlation such as key-key or query-query. (c) Residual Cross-correlation Self-attention (RCS) and Semantic Feedback Refinement (SFR). 
    (d) The performance comparison between our methods and baselines. 
    }
    \label{fig:method_simplify_all}
    \vspace{-4mm}
\end{figure*}
Recently, we have witnessed the unprecedented breakthrough of vision-language models (VLMs)~\cite{radford2021learning, caron2021emerging, darcet2023vision, oquab2023dinov2, kirillov2023segment}, which are de facto foundation models across various downstream tasks such as zero-shot classification~\cite{jia2021scaling, radford2021learning}, natural language processing~\cite{brown2020language, devlin2018bert, raffel2020exploring} and visual question answering~\cite{antol2015vqa, khan2022weakly, yu2022coca}. Especially, VLMs exhibit surprising open-vocabulary recognition capabilities because they are trained on large-scale image-text pairs with contrastive learning. Despite impressive performance achieved in image-level open-vocabulary tasks, they struggle with dense prediction tasks such as semantic segmentation due to the well-known limitation in localization ability~\cite{wang2025sclip, rao2022denseclip}.

To this end, some previous works~\cite{xu2022groupvit, luo2023segclip, xu2023learning, cha2023learning, zhang2023uncovering, ren2023viewco, xing2023rewrite} make great efforts to address these limitations by fine-tuning CLIP with pixel-level annotations. These methods~\cite{luo2023segclip, xu2023side, zhou2023zegclip} not only consume expensive annotation costs but also are easily biased toward the training data, eliminating the generalization ability within CLIP. This has motivated a growing interest in training-free methods~\cite{zhou2022extract, wang2025sclip, hajimiri2025naclip, lan2024clearclip, lan2024proxyclip, shao2025explore} that aim to adapt pre-trained representations of CLIP for semantic segmentation without additional training and maximumly maintain the generalization ability of CLIP simultaneously. These approaches typically attribute the inferior results of CLIP on dense prediction to the self-attention layer in the last block of CLIP (last col. in Fig.~\ref{fig:differ_layer_attention}), which presents spatial-invariant attention. For example, SCLIP proposes to replace original query-key self-attention with query-query and key-key self-attention. NACLIP~\cite{hajimiri2025naclip} employs key-key self-attention with neighbor priors to enhance attention across adjacent patches. We summarize these attention paradigms as self-correlation self-attention (SCSA, Fig.~\ref{fig:method_simplify_all} (b)), which achieves better spatial-covariant features than the original last self-attention layer (Fig.~\ref{fig:method_simplify_all} (a)). However, they overlook the cross-correlation self-attention (C$^2$SA) \footnote{For clarity, cross-correlation self-attention (C$^2$SA) is the standard self-attention that uses query and key to calculate attention. In contrast, self-correlation self-attention (SCSA) uses query or key pairs solely to calculate attention, such as key-key or query-query attention.} properties that capture the diverse and spatial correspondence. Ideally, if the attention is supervised by the pixel-level data, the C$^2$SA will learn densely class-specific representations. We, therefore, ask: \textit{Can we obtain such properties of the C$^2$SA from CLIP even if it does not receive pixel-level supervision?}

We answer this question by examining the C$^2$SA in all layers and reveal that the attention of C$^2$SA in non-last layers exhibits localization properties unlike that in the last layer. We visualize the attention of CLIP across different layers in Fig.~\ref{fig:differ_layer_attention}. We can observe that the attention in the last layer is spatial-invariant while the attentions in other layers exhibit class-specific features and localization properties. For example, the attention of the ``cat'' in the top image can attend to other ``cat'' regions in the image to some extent. This motivates us to propose a novel Residual Cross-correlation Self-attention (RCS) that borrows the cross-correlation self-attention from intermediate layers to make a residual connection with the attention in the last block. The cross-correlation could reorganize the spatial information thus unleashing the potential of CLIP for dense vision-language inference.
Moreover, to explicitly enhance the focus on regions of the same categories and local consistency, we also propose a Semantic Feedback Refinement (SFR) module to leverage the semantic segmentation map to further tweak the attention score. It is worth noting that our method termed \textbf{ResCLIP} (Fig.~\ref{fig:method_simplify_all} (c)) is orthogonal to existing works and can be seamlessly integrated into existing approaches as a plug-and-play module.

We conduct extensive experiments across eight segmentation benchmarks, demonstrating substantial potential of CLIP for open-vocabulary segmentation. As a plug-and-play solution, we integrate our method with three leading models: SCLIP, ClearCLIP, and NACLIP. The experimental results demonstrate significant improvements, with consistent mIoU gains across all datasets. As shown in Fig.~\ref{fig:method_simplify_all} (d), equipped with our method, the performance will be significantly boosted across all models. For example, our method makes improvements from $1.7\%$ to $13.1\%$ mIoU compared with counterparts. We also have conducted many ablation studies in both quantitative and qualitative ways to analyze the components of our method.

Our contributions can be summarized as follows:
\begin{itemize}
    \item We reveal that attention from intermediate layers of CLIP exhibits class-specific features and localization properties. To the best of our knowledge, this work is the first work to discover that the cross-correlation self-attention from intermediate layers present localization properties and can heal the attention in the last layer.
    \item We propose a novel training-free approach, terms ResCLIP, including Residual Cross-correlation Self-attention (RCS) and Semantic Feedback Refinement (SFR) modules. These two modules can rectify the attention in the last layer to capture class-specific features and local consistency so that improve CLIP model for dense vision-language prediction tasks.
    \item We have conducted extensive experiments on open-vocabulary semantic segmentation tasks including eight widely used benchmarks. The results in experiments demonstrate the effectiveness of our proposed method.
\end{itemize} 
\section{Related Work}
\label{sec:formatting}

\noindent\textbf{Vision-language Foundation Models.}
In recent years, VLMs have emerged as a general paradigm in various vision tasks, demonstrating remarkable capabilities in zero-shot and few-shot learning scenarios. Unlike Segment Anything Model (SAM)~\cite{kirillov2023segment, zou2024segment, yang2024tuning} which focus on promoting semantic-agnostic segmentation, a series of contrastive learning-based approaches~\cite{alayrac2020self, cherti2023reproducible, jia2021scaling, li2021align, miech2020end, radford2021learning, xu2023demystifying, yao2021filip, yuan2021florence} have shown exceptional adaptability across diverse downstream tasks. CLIP, in particular, achieves robust vision-text alignment through training on image-text pairs~\cite{fini2023improved, wu2023medklip, yang2022unified}. This has spawned numerous extensions to expand tasks such as visual question answering~\cite{khan2022weakly, kim2021vilt}, image captioning~\cite{li2022blip} and downstream inference capabilities~\cite{cherti2023reproducible, wang2022learning}.

\noindent\textbf{Open-vocabulary Semantic Segmentation (OVSS).} The OVSS extends traditional segmentation by enabling pixel-level dense prediction for arbitrary categories specified through text descriptions. Recent advances in vision-language models, particularly CLIP, have catalyzed significant progress~\cite{cherti2023reproducible, radford2021learning, xu2023demystifying} in this field. For their supervision requirements: fully-supervised methods~\cite{han2023open, jiao2023learning, liang2023open} that fine-tune CLIP using pixel-level annotations, weakly-supervised approaches~\cite{xu2022groupvit, xing2023rewrite, cha2023learning, luo2023segclip, xu2023learning, zhang2023uncovering, ren2023viewco, xing2023rewrite} that leverage image-text pairs for training, and training-free methods~\cite{zhou2022extract, li2023clip, wang2025sclip, hajimiri2025naclip, lan2024clearclip, kang2024defense, radford2021learning, shin2022reco, bousselham2024grounding} that modify architecture of CLIP with minimal changes. While fully-supervised methods achieve strong performance but require extensive labeled data, and weakly-supervised methods like GroupViT~\cite{xu2022groupvit} introduce specialized architectures with group tokens, training-free approaches such as MaskCLIP~\cite{zhou2022extract} focus on adapting self-attention mechanisms to enable dense prediction capabilities. However, these existing methods often struggle to fully utilize semantic understanding of CLIP due to limited fine-tuning datasets~\cite{sun2024clip} and their local understanding capability.

\noindent\textbf{Different Self-attention for Dense Visual Features.}
Recent training-free approaches show great interest in modifying attention mechanisms to enhance dense visual representation capabilities of CLIP. While vanilla CLIP employs query-key multiplication to obtain holistic visual representations that are invariant to spatial positions~\cite{zhou2024image}, subsequent works have explored various attention modifications. SCLIP~\cite{wang2025sclip} introduces self-correlation attention by combining query-query and key-key products as final layer attention weights to capture spatial-covariant features. This was followed by approaches such as GEM~\cite{bousselham2024grounding} presents a way to calculate the attention matrix as the combination of query-query, query-key and value-value attention, NACLIP~\cite{hajimiri2025naclip} emphasizes key-key products with Gaussian kernels, ClearCLIP~\cite{lan2024clearclip} utilizes query-query interactions, CLIPtrase~\cite{shao2025explore} tries to use weighted average of self-correlation attention to cluster the "global" patch for segmentation and ProxyCLIP~\cite{lan2024proxyclip} exploring attention combination with vision foundation models such as SAM. Different from these works, we discover that the cross-correlation self-attention from intermediate layers present localization properties and can heal the attention in the last layer. Therefore, we propose a novel training-free approach, terms ResCLIP, including Residual Cross-correlation Self-attention (RCS) and Semantic Feedback Refinement (SFR) modules. These two modules can rectify the attention in the last layer to capture class-specific features and local consistency so that improve CLIP model for dense vision-language prediction tasks.
\section{Methodology}
\label{sec:methodology}
In this section, we begin with an introduction to CLIP and its application to open-vocabulary semantic segmentation in a training-free paradigm in Sec.~\ref{sec:method_introduction}. Then, we describe the design of the proposed method including Residual Cross-correlation Self-attention (RCS) and Semantic Feedback Refinement (SFR) modules in Sec.~\ref{sec:our_method}.
\subsection{Preliminary}
\label{sec:method_introduction}

\noindent\textbf{Vision Encoder Architecture of CLIP.}
CLIP utilizes a Vision Transformer (ViT)~\cite{radford2021learning} to encode images into a representation aligning with textual descriptions. In conventional ViT~\cite{dosovitskiy2020image}, an input image $ H \times W \times 3 $ is partitioned into a grid of non-overlapping patches of size $ P \times P $. This results in $h=H/P$ rows, and $w=W/P$ columns of patches. Each patch is then being projected into a $d$ dimensional space, which can be presented by vectorized feature $x_{i} \in \mathbb{R}^d$ with preserved spatial relationships through explicit positional embeddings. Therefore, the input at each layer can be formulated as a sequence of visual tokens $X = \{ x_\text{cls}, x_1, ..., x_{h \times w} \} \in \mathbb{R}^{(1+hw) \times d}$, where $x_{\text{cls}}$ denotes the class token to capture global information. These visual tokens will feed into several multi-head self-attention layers to obtain the final token representations.

\noindent\textbf{Self-attention Module.}
The core of the transformer encoder is the self-attention mechanism (see Fig.~\ref{fig:method_simplify_all} (a)), which enables the model to capture relationships between different patches. Note that we only consider the single-head self-attention for simplicity. The self-attention is given by:
\begin{equation}
    q,k,v= \text{Proj}_{q,k,v}(\text{LN}(X)), \label{eq:qkv}
\end{equation}
\begin{equation}
    S_{qk}= qk^T / \sqrt{d_{k}}, \label{eq:sim}
\end{equation}
\begin{equation}
    \text{Attn}(S_{qk})=\text{softmax}(S_{qk})=\text{softmax}(qk^T / \sqrt{d_{k}}), \label{eq:attn}
\end{equation}
where $q, k, v$ indicates the query, key and value, respectively. The $\text{LN}$ is layer normalization, $\text{Proj}$ denotes a projection layer, and $d_k$ is the dimension of key $k$. The attention scores $S_{qk} \in \mathbb{R}^{(1+hw) \times (1+hw)}$ imply the intrinsic global structural dependencies between patches, from which the attention map $\text{Attn}(S_{qk})$ is obtained through the softmax normalization.

\noindent\textbf{Dense Vision-language Inference.}
CLIP was originally trained on large-scale image-text pairs using contrastive loss, demonstrating promising results in open-vocabulary image recognition tasks. The text representations $X_{\text{text}}=\{t_1, t_2, ..., t_c\}$ are obtained through a text encoder and are used to align with $cls$ token $x_{\text{cls}}$ of visual features for each image-text pair. A natural idea is to extend the CLIP model to dense vision-language inference tasks such as open-vocabulary semantic segmentation by calculating the similarity between the dense visual tokens $X_{\text{dense}} = \{ x_1, ..., x_{h \times w} \} \in \mathbb{R}^{hw \times d}$ and text tokens  $X_{\text{text}}$. In particular, we can obtain the segmentation map as follows,
\begin{equation}
\mathcal{M}= \mathop{\arg\max} \cos (X_{\text{dense}},X_{\text{text}}).
\label{eq:seg_mask}
\end{equation}
However, the obtained semantic segmentation maps are full of noise due to the deficiency in localization ability in the CLIP model. The purpose of this work is to improve the performance of dense vision-language inference in a training-free manner so that maximumly reserves the generalization ability of CLIP by adjusting the attention in the last layers. 

\begin{figure}
    \centering
    \includegraphics[width=\linewidth]{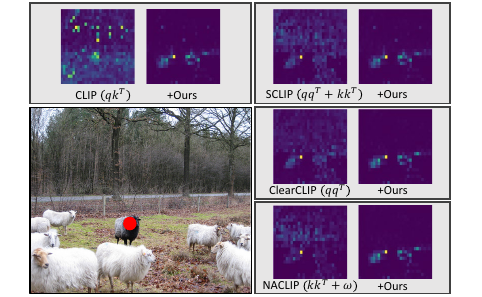}
    \vspace{-3mm}
    \caption{Comparison of attention maps across different versions of CLIP and ours.}
    \label{fig:ours_attn_motivation}
    \vspace{-4mm}
\end{figure}

\subsection{ResCLIP}
\label{sec:our_method}

As demonstrated in the previous works~\cite{zhou2022extract}, CLIP exhibits inherent limitations in pixel-level semantic segmentation. To address these challenges, recent studies~\cite{wang2025sclip, zhou2022extract, hajimiri2025naclip, lan2024clearclip} have proposed several approaches in a training-free manner and involve minimal modifications to the original model. These methods reveal that the attention in the last layer of CLIP exhibits spatial-invariant features, \ie, local features tend to be invariant to their spatial position in the image. Therefore, these methods~\cite{wang2025sclip,hajimiri2025naclip,lan2024clearclip} focus on reformulating the self-attention module in the final layer of CLIP by introducing a self-correlation self-attention (SCSA) to obtain the spatial-covariant features. These SCSA-based methods encourage each local token to attend to itself and the positions sharing similar features with it so that eliminates the spatial-invariant issues in the original CLIP, leading to improved dense prediction performance. 

However, the SCSA enforces the self-similar features response and lacks the capacity to capture the cross-feature dynamics that cross-correlation self-attention (C$^2$SA) provides. The cross-correlation self-attention typically captures the diverse and spatial correspondence, which is beneficial for localization. But the C$^2$SA in the last layer has shown spatial-invariant properties in CLIP. Through careful experiments, we reveal that the self-attention in non-last layers of CLIP exhibits class-specific features and localization properties (See Fig.~\ref{fig:differ_layer_attention}). For example, the attention of the ``cat'' in the top image can attend to other ``cat'' regions in the image to some extent.

\begin{figure*}
    \centering
    \includegraphics[width=0.9\linewidth]{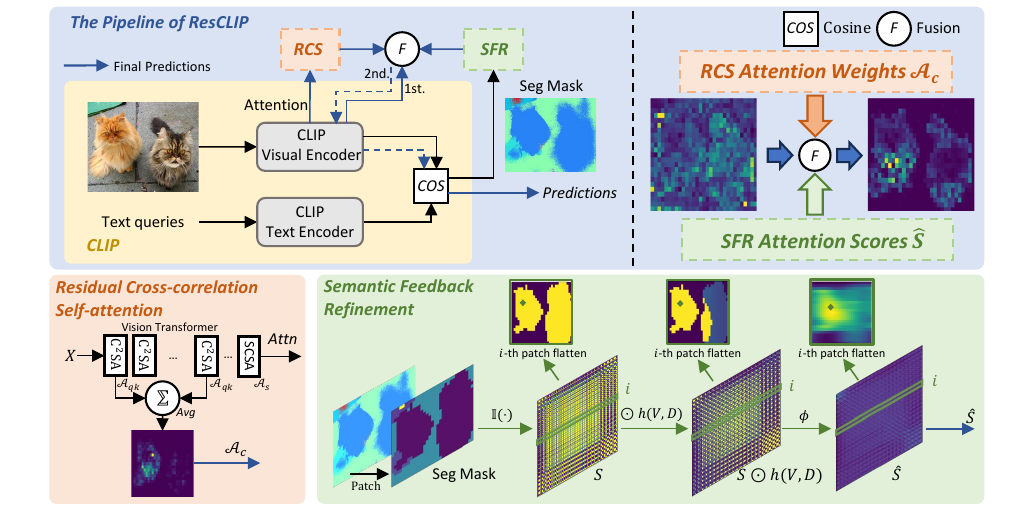}
    \vspace{-3mm}
    \caption{Overview of our ResCLIP consisting of Residual Cross-correlation Self-attention (RCS) and Semantic Feedback Refinement (SFR). The RCS module enhances attention mechanism of CLIP by fusing C$^2$SA from non-last layers $\mathcal{A}_c$ with SCSA $\mathcal{A}_s$ to capture richer spatial information. The SFR module leverages an initial segmentation mask (black arrows) to refine attention scores. These refined attention scores $\hat{S}$ are combined with RCS to adjust the attention in the last layer of CLIP and produce the final prediction (blue arrows).
    }
    \label{fig:pipeline_all}
    \vspace{-4mm}
\end{figure*}

Therefore, we propose a novel training-free approach that aggregates the C$^2$SA from intermediate layers of CLIP to remold the attention in the last block. The cross-correlation could reorganize the spatial information thus unleashing the potential of CLIP for dense vision-language inference. In particular, we propose a novel Residual Cross-correlation Self-attention (RCS) that aggregates the C$^2$SA from intermediate layers to make a residual connection with the attention in the last block. The RCS could reorganize the spatial information thus unleashing the potential of CLIP for dense vision-language inference. Moreover, to explicitly enhance the focus on regions of the same categories and local consistency, we also propose a Semantic Feedback Refinement (SFR) module to leverage the semantic segmentation map to further tweak the attention score. As shown in Fig.~\ref{fig:ours_attn_motivation}, our method could improve the attention of the baseline methods to attend more semantic-related regions. The overview of our method is present in Fig.~\ref{fig:pipeline_all}.

\subsubsection{Residual Cross-correlation Self-attention}
\label{sec:RCS}
The Residual Cross-correlation Self-attention (RCS) module borrows the attention from the intermediate layers to remold the attention module in the last layer. Specifically, we first extract the C$^2$SA from intermediate layers and aggregate information from them through the average operation. Formally, we denote the C$^2$SA attention $\text{Attn}^{i}(S_{qk})$ as $\mathcal{A}_{qk}^i$ in $i$-th layer, the aggregated attention $\mathcal{A}_{c}$ can be calculated as follows,
\begin{equation}
    \label{eq:avg_qk}
    \mathcal{A}_{c} = \frac{1}{N} \sum _{i=s}^{e} \mathcal{A}_{qk}^i,
\end{equation}
where $N=e-s+1$ indicates the number of layers from the start layer $s$ to the end layer $e$. Based on this aggregated attention, the RCS attention can be formulated as follows,
\begin{equation}
    \label{eq:rescs}
    \mathcal{A}_{rcs} = (1 - \lambda_{rcs}) \cdot \mathcal{A}_{s} + \lambda_{rcs} \cdot \mathcal{A}_{c},
\end{equation}
where $\mathcal{A}_{s}$ is the SCSA attention used in previous works~\cite{wang2025sclip,lan2024clearclip} and $\lambda_{rcs}$ is the trade-off parameter. Our RCS absorbs information from both the $\mathcal{A}_{s}$ and $\mathcal{A}_{c}$, which capture the local patch structure and cross-feature correspondence. This attention could better reorganize the information in the last layer and improve the final classification. 

\subsubsection{Semantic Feedback Refinement}
\label{sec:SFR}
\begin{table*}[!t]
\centering
\caption{
Open-vocabulary semantic segmentation quantitative comparison on datasets \textit{without} a background class. Our results are marked in \colorbox{Gray!16}{gray}. The best results on each dataset for different encoders are \textbf{bolded}.
}
\vspace{-3mm}
  \tabcolsep3pt
  \centering
  \resizebox{0.8\textwidth}{!}{%
  \begin{tabular}{lcc|cccccc}
    \toprule
    Methods & \makecell[c]{\textit{Training-free?}} & Encoder & VOC20 & Context59 & Stuff & Cityscape & ADE20k & Avg. \\
    \hline
    GroupViT \cite{xu2022groupvit} & $\times$ & ViT-S/16 & 79.7 & 23.4 & 15.3 & 11.1 & 9.2 & 27.7\\
    CoCu \cite{xing2023rewrite} & $\times$ & ViT-S/16 & - & - & 13.6 & 15.0 & 11.1 & - \\
    TCL \cite{cha2023learning} & $\times$ & ViT-B/16 & 77.5 & 30.3 & 19.6 & 23.1 & 14.9 & 33.1\\
    CLIP \cite{radford2021learning} & \checkmark & ViT-B/16 & 41.8 & 9.2 & 4.4 & 5.5 & 2.1 & 12.6\\
    MaskCLIP \cite{zhou2022extract} & \checkmark & ViT-B/16 & 74.9 & 26.4 & 16.4 & 12.6 & 9.8 & 28.0\\
    ReCo \cite{shin2022reco} & \checkmark & ViT-B/16 & 57.7 & 22.3 & 14.8 & 21.1 & 11.2 & 25.4\\
    CLIPSurgery \cite{li2023clip} & \checkmark & ViT-B/16 & - & - & 21.9 & 31.4 & - & - \\
    \hline
    SCLIP \cite{wang2025sclip} & \checkmark & ViT-B/16 & 80.4 & 34.2 & 22.4 & 32.2 & 16.1 & 37.1\\
    \rowcolor{Gray!16}
    +ResCLIP(ours) & \checkmark & ViT-B/16 & 84.6 & 35.8 & 23.9 & 34.4 & 17.6 & 39.3
    \hfill\textcolor{ForestGreen}{(+2.2)}\\
 
    ClearCLIP \cite{lan2024clearclip} & \checkmark & ViT-B/16 & 80.9 & 35.9 & 23.9 & 30.0 & 16.7 & 37.5\\
    \rowcolor{Gray!16}
    +ResCLIP(ours) & \checkmark & ViT-B/16 & \textbf{87.1} & 36.4 & 24.3 & 34.5 & 17.8 & 40.0
    \hfill\textcolor{ForestGreen}{(+2.5)}\\

    NACLIP \cite{hajimiri2025naclip} & \checkmark & ViT-B/16 & 79.7 & 35.2 & 23.3 & 35.5 & 17.4 & 38.2\\
    \rowcolor{Gray!16}
    +ResCLIP(ours) & \checkmark & ViT-B/16 & 86.0 & \textbf{36.8} & \textbf{24.7} & \textbf{35.9} & \textbf{18.0} & \textbf{40.3}
    \hfill\textcolor{ForestGreen}{(+2.1)}\\

    \hline
    CLIP \cite{radford2021learning} & \checkmark & ViT-L/14 & 15.8 & 4.5 & 2.4 & 2.9 & 1.2 & 5.4 \\
    MaskCLIP \cite{zhou2022extract} & \checkmark & ViT-L/14 & 30.1 & 12.6 & 8.9 & 10.1 & 6.9 & 13.7 \\
    
    SCLIP \cite{wang2025sclip} & \checkmark & ViT-L/14 & 60.3 & 20.5 & 13.1 & 17.0 & 7.1 & 23.6 \\
    \rowcolor{Gray!16} 
    +ResCLIP(ours) & \checkmark & ViT-L/14 & 83.9 & 30.6 & 21.2 & 32.2 & 15.8 & 36.7
    \hfill\textcolor{ForestGreen}{(+13.1)}\\
    
    ClearCLIP \cite{lan2024clearclip} & \checkmark & ViT-L/14 & 80.0 & 29.6 & 19.9 & 27.9 & 15.0 & 34.5\\
    \rowcolor{Gray!16} 
    +ResCLIP(ours) & \checkmark & ViT-L/14 & 84.2 & 33.4 & 22.3 & \textbf{34.1} & 17.9 & 38.4
    \hfill\textcolor{ForestGreen}{(+3.9)}\\

    NACLIP \cite{hajimiri2025naclip} & \checkmark & ViT-L/14 & 78.7 & 32.1 & 21.4 & 31.4 & 17.3 & 36.2 \\
    \rowcolor{Gray!16} 
    +ResCLIP(ours) & \checkmark & ViT-L/14 & \textbf{85.5} & \textbf{34.5} & \textbf{23.4} & 33.7 & \textbf{18.2} & \textbf{39.1}
    \hfill\textcolor{ForestGreen}{(+2.9)}\\
   
  \bottomrule
  \label{tab:results_wo_background}
  \end{tabular} 
  }
  \vspace{-5mm}
\end{table*}
Although our RCS module not only provides spatial-invariant attention but also cross-feature dynamics, the attention is still not real attention under the pixel level supervision. Some works such as NACLIP~\cite{hajimiri2025naclip} add prior regularization to the attention by paying attention to the neighbor patches with discretized Gaussian kernels into attention maps. However, these kernels are isotropic and thus may introduce extra context due to the shapes of the objects being versatile. Besides, ideal attention should attend to the objects that share the same categories~\cite{oquab2023dinov2}. To this end, we propose a Semantic Feedback Refinement (SFR) module that improves the attention with the same semantics and maintains the locality simultaneously by using the semantic segmentation map from the CLIP model as feedback. 

In particular, we begin with a semantic segmentation map, which is obtained by our RCS module, and then use this semantic segmentation map to further tweak the attention in the last layer. This loop would obtain better attention, resulting in improved performance. Formally, suppose that we have an attention score $S \in \mathbb{R}^{{hw} \times {hw}}$, each element $S_{ij}$ in $S$ indicates the attention from $i$-th patch to $j$-th patch. Now, we consider there is a semantic segmentation map $\mathcal{M} \in \mathbb{R}^{hw}$. For brevity, we only consider the attention score $S^i \in \mathbb{R}^{hw}$ for one patch as an example, it can be obtained as follows,
\begin{equation}
    \label{eq:mask_create}
    S^i_{m,n} = \mathbb{I}(\mathcal{M}_{i^\prime, j^\prime} == \mathcal{M}_{m,n}),
\end{equation}
where $m \in \{1,2,...,h\}$, $n \in \{1,2,...,w\}$, $i^\prime=\left\lfloor i/w \right\rfloor$ and $j^\prime=(i \mod w)$ indicates the grid index in mask $\mathcal{M}$ for the $i$-row attention $S^i$. And the $\mathbb{I}(\cdot)$ is the indicator function where 1 when $\cdot$ is true, otherwise 0. 

To further maintain the locality, we emphasize the attention to the adjacent patches of the same class. We first identify connected and disjoint regions, then reduce the attention to the disjoint ones. To be specific, we first define a decay function as follows,
\begin{equation}
    \label{eq:decay_func}
    h(V,D) = V + (1-V) \cdot D,
\end{equation}
where $V$ is the mask where $V_{mn}=1$ if there a valid path from $(m, n)$ to $(i^\prime, j^\prime)$, otherwise 0. $D$ is a distance-based decay function, which is defined as:
\begin{align}
D(p,q)=exp( - \frac{d(p,q)}{max(d(\cdot,\cdot))}),
\label{eq:filter_d_decay}
\end{align}
where $p$ and $q$ denote the grid coordinates for the patches. We use the Chebyshev Distance for simplicity which measures the most significant difference with different patches as follows,
\begin{align}
d(p,q)=\max (|p_x-q_x|,|p_y-q_y|).
\label{eq:chebyshev_distance}
\end{align}

Therefore, we can apply the decay function to attention score $S^i$ as follows,
\begin{align}
\hat{S}^{i} = \phi (S^i \odot h(V, D)), \label{eq:srec_before_gauss}
\end{align}
where $\phi$ is a conventional 1-dimensional Gaussian kernel to smooth the final attention score.
This smoothing operation enhances the generalization of attention scores while preserving the row-independence property of the original attention mechanism. By employing the above operation to each row, we can obtain the entire attention score map $\hat{S}$. Finally, we combine the original attention scores with our semantic feedback refinement score, which can be formulated as follows,
\begin{align}
S_{r} = (1 - \lambda_{sfr}) \cdot S_{s} + \lambda_{sfr} \cdot \hat{S},
\label{eq:sfr}
\end{align}
where $S_{s}$ is the attention score of SCSA and $\lambda_{sfr}$ is the trade-off parameter. Now, we lead to the final version of ResCLIP, the final attention can be given by:
\begin{equation}
    \label{eq:all_final}
    \mathcal{A}_{ResCLIP} = (1 - \lambda_{rcs}) \cdot \mathcal{A}_{sfr} + \lambda_{rcs} \cdot \mathcal{A}_{c}, 
\end{equation}
where $\mathcal{A}_{sfr}=softmax(S_r)$. To this end, residual attention $\mathcal{A}_{ResCLIP}$ benefit from 1) SCSA attention that provides spatial-covariant features as mentioned in previous works~\cite{wang2025sclip,hajimiri2025naclip}; 2) C$^2$SA attention from intermediate layers that capture the rich spatial correspondence; 3) SFR attention that explicitly enhances
the focus on regions of the same categories and local consistency.
\section{Experiments}
\label{sec:experiments}

\begin{table*}[!t]
  \tabcolsep8pt
  \centering
  \caption{Open-vocabulary semantic segmentation quantitative comparison on datasets \textit{with} a background class. Our results are marked in \colorbox{Gray!16}{gray}. The best results on each dataset for different encoders are \textbf{bolded}.
  }
  \vspace{-3mm}
  \resizebox{0.75\textwidth}{!}{
  \begin{tabular}{lcc|cccc}
    \toprule
    Methods & \textit{Training-free?} & Encoder & VOC21 & Context60 & Object & Avg. \\
    
    \hline
    GroupViT \cite{xu2022groupvit} & $\times$ & ViT-S/16 & 50.4 & 18.7 & 27.5 & 32.2 \\
    SegCLIP \cite{luo2023segclip} & $\times$ & ViT-S/16 & 52.6 & 24.7 & 26.5 & 34.6 \\
    OVSegmentor \cite{xu2023learning} & $\times$ & ViT-B/16 & 53.8 & 20.4 & 25.1 & 33.1 \\
    PGSeg \cite{zhang2023uncovering} & $\times$ & ViT-S/16 & 53.2 & 23.8 & 28.7 & 35.2 \\
    ViewCo \cite{ren2023viewco} & $\times$ & ViT-S/16 & 52.4 & 23.0 & 23.5 & 33.0\\
    CoCu \cite{xing2023rewrite} & $\times$ & ViT-S/16 & 40.9 & 21.2 & 20.3 & 27.5\\
    TCL \cite{cha2023learning} & $\times$ & ViT-B/16 & 51.2 & 24.3 & 30.4 & 35.3 \\
    
    CLIP \cite{radford2021learning} & \checkmark & ViT-B/16 & 16.2 & 7.7 & 5.5 & 9.8 \\
    MaskCLIP \cite{zhou2022extract} & \checkmark & ViT-B/16 &  38.8 & 23.6 & 20.6 & 27.7 \\
    ReCo \cite{shin2022reco} & \checkmark & ViT-B/16 & 25.1 & 19.9 & 15.7 & 20.2 \\
    CLIPSurgery \cite{li2023clip} & \checkmark & ViT-B/16 & - & 29.3 & - & - \\
    GEM \cite{bousselham2024grounding} & \checkmark & ViT-B/16 & 46.2 & 32.6 & - & - \\
    
    \hline
    SCLIP \cite{wang2025sclip} & \checkmark & ViT-B/16 & 59.1 & 30.4 & 30.5 & 40.0 \\
    \rowcolor{Gray!16} 
    +ResCLIP(ours) & \checkmark & ViT-B/16 & 60.7 & 32.9 & 34.3 & 42.7 
    \hfill\textcolor{ForestGreen}{(+2.7)}\\
         
    ClearCLIP \cite{lan2024clearclip} & \checkmark & ViT-B/16 & 51.8 & 32.6 & 33.0 & 39.1 \\
    \rowcolor{Gray!16} 
    +ResCLIP(ours) & \checkmark & ViT-B/16 & 59.0 & 32.9 & 34.0 & 42.0 
    \hfill\textcolor{ForestGreen}{(+2.9)}\\

    NACLIP \cite{hajimiri2025naclip} & \checkmark & ViT-B/16 & 58.9 & 32.2 & 33.2 & 41.4 \\
    \rowcolor{Gray!16} 
    +ResCLIP(ours) & \checkmark & ViT-B/16 & \textbf{61.1} & \textbf{33.5} & \textbf{35.0} & \textbf{43.2} 
    \hfill\textcolor{ForestGreen}{(+1.8)}\\
    
    \hline
    SCLIP \cite{wang2025sclip} & \checkmark & ViT-L/14 & 44.4 & 22.3 & 24.9 & 30.5 \\
    \rowcolor{Gray!16} 
    +ResCLIP(ours) & \checkmark & ViT-L/14 & 52.8 & 28.7 & 30.9 & 37.4
    \hfill\textcolor{ForestGreen}{(+6.9)}\\
    
    ClearCLIP \cite{lan2024clearclip} & \checkmark & ViT-L/14 & 48.7 & 28.3 & 29.7 & 35.5\\
    \rowcolor{Gray!16} 
    +ResCLIP(ours) & \checkmark & ViT-L/14 & 50.7 & 29.8 & 31.1 & 37.2
    \hfill\textcolor{ForestGreen}{(+1.7)}\\

    NACLIP \cite{hajimiri2025naclip} & \checkmark & ViT-L/14 & 52.2 & 28.7 & 29.9 & 36.9 \\
    \rowcolor{Gray!16} 
    +ResCLIP(ours) & \checkmark & ViT-L/14 & \textbf{54.1} & \textbf{30.9} & \textbf{32.5} & \textbf{39.2}
    \hfill\textcolor{ForestGreen}{(+2.3)}\\
    
  \bottomrule
  \label{tab:results_w_background}
  \end{tabular}
  }
  \vspace{-5mm}
\end{table*}
\subsection{Experimental Setups}
\label{sec:Experimental Setups}
\noindent\textbf{Datasets.}
We conduct comprehensive evaluations on eight widely-adopted benchmark datasets for open-vocabulary semantic segmentation. Following prior works~\cite{wang2025sclip, hajimiri2025naclip, lan2024clearclip}, these datasets can be categorized into two groups based on the presence of a background category, whose names are abbreviated in parentheses for brevity. Firstly, datasets with background category: PASCAL VOC 2012~\cite{everingham2015pascal} (VOC21), PASCAL Context~\cite{mottaghi2014role} (Context60) and COCO Object~\cite{caesar2018coco} (Object). Secondly, datasets with background category: COCO-Stuff~\cite{caesar2018coco} (Stuff), Cityscapes~\cite{cordts2016cityscapes} and ADE20K-150~\cite{zhou2019semantic}. Additionally, we follow the construction of removing the background class in PASCAL VOC20~\cite{everingham2015pascal} (VOC20) and PASCAL Context59~\cite{mottaghi2014role} (Context59). Specifically, input images are resized to have a shorter side of 336 pixels, except for Cityscapes, where we use 560 pixels due to its inherently high-resolution images. We perform slide inference using a 224×224 window with a stride of 112 following~\cite{barsellotti2024fossil, cha2023learning, li2023clip, wang2025sclip, xu2022groupvit, hajimiri2025naclip, lan2024clearclip}.

\noindent\textbf{Baselines.}
We compare our work with a comprehensive range of OVSS methods, including direct baseline CLIP~\cite{radford2021learning}, previous state-of-the-art training-free approaches: MaskCLIP~\cite{zhou2022extract}, ReCo~\cite{shin2022reco}, CLIPSurgery~\cite{li2023clip}, GEM~\cite{bousselham2024grounding}, SCLIP~\cite{wang2025sclip}, NACLIP~\cite{hajimiri2025naclip} and ClearCLIP~\cite{lan2024clearclip}. We also include a few influential weakly supervised methods, such as GroupViT~\cite{xu2022groupvit}, CoCu~\cite{xing2023rewrite}, TCL~\cite{cha2023learning}, SegCLIP~\cite{luo2023segclip}, OVSegmentor~\cite{xu2023learning}, PGSeg~\cite{zhang2023uncovering}, and ViewCo~\cite{ren2023viewco}. Unless explicitly mentioned, all reported results are from the respective papers. 
Besides, we select recent state-of-the-art methods SCLIP~\cite{wang2025sclip}, NACLIP~\cite{hajimiri2025naclip}, ClearCLIP~\cite{lan2024clearclip} with specialized attention designs as baselines and evaluate performance of our method when integrated with these approaches. 
We also present results based on the ViT-L/14 backbone for a comprehensive evaluation.

\noindent\textbf{Implementation details.}
In our experiments, we utilize the implementations provided by MMSegmentation~\cite{contributors2020mmsegmentation}. Following TCL~\cite{cha2023learning}, we abstain from computationally intensive post-processing techniques that could lead to unfair comparisons, such as PAMR~\cite{araslanov2020single} (used in TCL~\cite{cha2023learning}, NACLIP~\cite{hajimiri2025naclip}) and DenseCRF~\cite{krahenbuhl2011efficient} (used in ReCo~\cite{shin2022reco}). We employ only standard ImageNet prompts~\cite{radford2021learning} without additional textual prompting strategies. By default, our comparative experiments are conducted using the ViT-B/16~\cite{dosovitskiy2020image} backbone, with ablation studies and sensitivity analyses performed on top of NACLIP. Our approach operates in a fully training-free manner, requiring neither retraining nor fine-tuning. We evaluate all semantic segmentation tasks using the mean Intersection over Union (mIoU) metric.

\subsection{Main Results}
\label{sec:main_results}
\noindent\textbf{Quantitative results.}
Table~\ref{tab:results_wo_background} summarizes the performance of various open-vocabulary semantic segmentation models on datasets without a background class. Our ResCLIP demonstrates significant improvements when integrated with state-of-the-art approaches including SCLIP~\cite{wang2025sclip}, ClearCLIP~\cite{lan2024clearclip}, and NACLIP~\cite{hajimiri2025naclip}. Notably, when combined with NACLIP~\cite{hajimiri2025naclip}, our method achieves state-of-the-art performance, surpassing mainstream weakly supervised approaches. As a plug-and-play solution, consistent improvements are observed across all five datasets compared to the baseline methods, showing the great potential of ResCLIP. We also evaluate performance on ViT-L/14 and observe that when adapting to different backbones, existing methods typically decrease by over $2\%$ mIoU, \eg, SCLIP~\cite{wang2025sclip} has a particularly severe drop of $13.5\%$ mIoU. However, with the integration of our method, this performance degradation is significantly mitigated, demonstrating the effectiveness of the proposed method.
Additionally, we conduct experiments on datasets with a background class in Table~\ref{tab:results_w_background}. We can observe that ResCLIP shows consistent improvements across all datasets with a background class over all the counterparts. Specifically, our method surpasses supervised approaches such as GroupViT~\cite{xu2022groupvit}, achieving state-of-the-art performance of $43.2\%$ mIoU when integrated with NACLIP~\cite{hajimiri2025naclip} on ViT-B/16 backbone. Regarding results of the VIT-L/14, our method still provides substantial improvements compared with other baselines. These results validate that the method effectively reduces noise of original attention in the last layer.

\begin{figure}
    \centering
    \includegraphics[width=\linewidth]{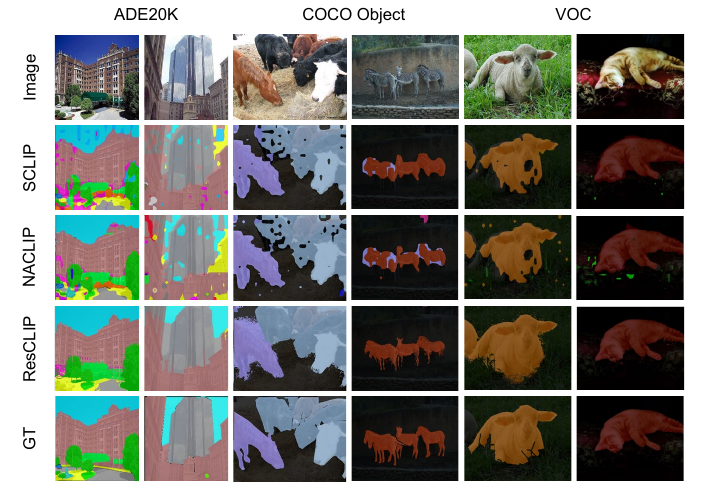}
    \vspace{-4mm}
    \caption{Qualitative comparison between different CLIP-based training-free segmentation methods.
    }
    \label{fig:experiment_visual}
    \vspace{-3mm}
\end{figure}
\noindent\textbf{Qualitative results.}
In Fig.~\ref{fig:experiment_visual}, we present qualitative comparison results with training-free methods such as SCLIP~\cite{wang2025sclip}, NACLIP~\cite{hajimiri2025naclip}, and ClearCLIP~\cite{lan2024clearclip}. From Fig.~\ref{fig:experiment_visual}, our ResCLIP predicts higher quality and more accurate segmentation maps with reduced noise. ResCLIP demonstrates superior attention to the internal regions of an object, avoiding hollow dense predictions in central areas (\eg, the 1st col. in VOC and the 1st col. in COCO Object). Compared to NACLIP, which employs key-key attention in the last layer, ResCLIP achieves both effective noise reduction and clearer segmentation maps (\eg, on ADE20k), demonstrating that ResCLIP effectively captures rich spatial correspondences between objects of the same category, while SFR successfully enhances local consistency. More qualitative results can be referred to our Supplementary.

\subsection{Experimental Analysis}
\label{sec:Experimental Analysis}

\begin{table}[t]
    \centering
    \caption{Ablation studies for the proposed method. 
    }
    \vspace{-3mm}
    \tablestyle{3.6pt}{1.0}
    \begin{tabular}{c|cc|cc}
    \toprule
        \multirow{2}{*}{Methods} & \multicolumn{2}{c|}{Module} & \multirow{2}{*}{\makecell[c]{mIoU}} & \multirow{2}{*}{\makecell[c]{$\Delta$}} \\ 
         &  \textit{RCS} & \textit{SFR} \\
         \hline
        Baseline (NACLIP~\cite{hajimiri2025naclip}) & - & - & 79.7 & - \\
        \hline
        \multirow{3}{*}{\makecell[c]{ResCLIP(Ours)}}
        &  \checkmark & & 85.5 & \textcolor{ForestGreen}{+5.8}\\
        &  & \checkmark & 81.5 & \textcolor{ForestGreen}{+1.8}\\
         & \cellcolor{Gray!16}\checkmark &
        \cellcolor{Gray!16}\checkmark & \cellcolor{Gray!16}\textbf{86.0} & \cellcolor{Gray!16}\textcolor{ForestGreen}{+\textbf{6.3}}\\
        \bottomrule
    \end{tabular}
    \label{tab:module_ablation}
    \vspace{-5mm}
\end{table}
\noindent\textbf{Ablation studies.}
We conduct ablation studies using NACLIP with ViT-B/16 backbone, as shown in Table~\ref{tab:module_ablation}. The RCS module alone improves mIoU from $79.7\%$ to $85.5\%$ on VOC20, demonstrating effectiveness of incorporating non-last layers $\mathcal{A}_{qk}$ information into the final attention map. The SFR module independently achieves a $1.8\%$ mIoU improvement, validating its effectiveness in enhancing attention between semantically similar regions while preserving local spatial consistency. When combined, these modules achieve a substantial $6.3\%$ mIoU improvement, reaching a mIoU of $86.0\%$, demonstrating their complementary utility.

\begin{figure}
    \centering
    \includegraphics[width=0.9\linewidth]{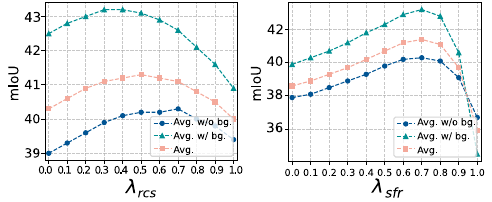}
    \vspace{-3mm}
    \caption{Analysis of hyper-parameters $\lambda_{rcs}$ and $\lambda_{sfr}$.
    }
    \label{fig:line_plot_convex_sensitive}
    \vspace{-2mm}
\end{figure}
\noindent\textbf{Sensitivity analysis of hyper-parameters.}
In Fig.~\ref{fig:line_plot_convex_sensitive}, we analyze sensitivity of two key trade-off hyper-parameters $\lambda_{rcs}$ and $\lambda_{sfr}$. 
For $\lambda_{rcs}$, optimal performance is achieved when $\lambda_{rcs}=0.5$. The parameter $\lambda_{sfr}$ demonstrates best results at $0.7$, with stable performance in the range from $0.6$ to $0.8$. Both parameters exhibit similar trends: moderate values enhance performance while extreme values (\eg, $1.0$) lead to degradation. This consistent behavior across different datasets suggests robustness of ResCLIP, requiring only coarse hyper-parameter tuning for effective deployment. 

\begin{table}[t]
    \centering
    \caption{Analysis of layer fusion strategies in our method.
    }
    \vspace{-4mm}
    \resizebox{0.38\textwidth}{!}{
    \begin{tabular}{cc|cc}
    \toprule
        \multicolumn{2}{c|}{\makecell[c]{Fusion Ways}}& \multirow{2}{*}{VOC20} & \multirow{2}{*}{Object} \\
        Methods & Layers & \\
        \hline
        \multirow{5}{*}{\makecell[c]{\textit{Cumulative Layers} \\ \textit{Aggregation}}} 
         & 1 $\rightarrow$ 1 & 81.1 & 34.0 \\
         & 1 $\rightarrow$ 3 & 82.6 & 34.8 \\ 
         & 1 $\rightarrow$ 5 & 84.1 & 34.8 \\
         & 1 $\rightarrow$ 7 & 84.8 & 34.9 \\ 
         & 1 $\rightarrow$ 9 & \textbf{85.5} & \textbf{35.0} \\ 
        \cline{1-4}
         \multirow{5}{*}{\makecell[c]{\textit{Sliding Window} \\ \textit{Aggregation}}} 
         & 2 $\rightarrow$ 5 & 84.2 & 34.8 \\
         & 4 $\rightarrow$ 7 & 85.0 & 34.7 \\
         & 6 $\rightarrow$ 9 & \textbf{86.0} & \textbf{35.0} \\
         & 8 $\rightarrow$ 11 & 85.8 & 34.8 \\
        \bottomrule
    \end{tabular}                       
    }
    \vspace{-6mm}
    \label{tab:fus_layer}
\end{table}
\noindent\textbf{Analysis of layer fusion strategies in RCS.}
As Eq.~(\ref{eq:avg_qk}) shows, our RCS method averages attention from some layers. We explore two strategies for the aggregation ($\mathcal{A}_{qk}$) from non-last layers to capture the rich spatial correspondence. In Table~\ref{tab:fus_layer}, we compare aggregation methods: Cumulative Layers Aggregation (CLA) and Sliding Window Aggregation (SWA). CLA starts from the first layer to the $n$-th while SWA applies a sliding window to aggregate attention where the window size is set to $4$. From Table~\ref{tab:fus_layer}, we can observe that 1) for the CLA, the model increases the performance with more layers involved. 2) when leveraging the sliding window, the $6\rightarrow9$ achieves the best results.
\section{Conclusion}
\label{sec:conclusion}
In this paper, we introduce ResCLIP, a novel framework that enhances ability of CLIP for dense vision-language inference in a training-free manner. We discovered that attention from intermediate layers of CLIP exhibits class-specific features and localization properties. We propose two modules: Residual Cross-correlation Self-attention (RCS) and Semantic Feedback Refinement (SFR). These two modules not only provide rich spatial correspondence but also enhance the focus on the regions with the same semantics and local consistency to remold the attention in the last layer of CLIP. Extensive experiments show that ResCLIP improves performance across various benchmarks, demonstrating the effectiveness of the proposed method.

\section*{Acknowledgments}
This work is supported by National Natural Science Foundation of China (No. 62176047), Sichuan Science and Technology Program (No. 2022YFS0600), National Natural Science Foundation of China (No. 62476051) and Sichuan Natural Science Foundation (No. 2024NSFTD0041).

{
    \small
    \bibliographystyle{ieeenat_fullname}
    \bibliography{main}
}

\clearpage
\setcounter{page}{1}

\renewcommand{\thesection}{\Alph{section}}
\renewcommand{\thesubsection}{\thesection.\arabic{subsection}}

\renewcommand{\thefigure}{A\arabic{figure}}
\renewcommand{\thetable}{A\arabic{table}}
\renewcommand{\theequation}{A\arabic{equation}}

\setcounter{equation}{0}
\setcounter{figure}{0}
\setcounter{table}{0}
\setcounter{section}{0}

\maketitlesupplementary

In this supplementary document, we present additional materials not included in the main manuscript due to page limitations. The supplementary content is outlined: 
\begin{itemize}
    \item Sec.~\ref{sec:supp_attention_comparison}: Additional attention comparison between the proposed method and previous works.
    \item Sec.~\ref{sec:supp_ablation_studies_on_backbones}: Ablation studies on different ViT backbones.
    \item Sec.~\ref{sec:supp_extension_on_other_VLMs}: Extension on other CLIP-like models.
    \item Sec.~\ref{sec:supp_infer_efficiency}: Inference efficiency analysis of designed models.
    \item Sec.~\ref{sec:supp_seg_visual}: More segmentation visualization results.
\end{itemize}

Now, we will present these materials as follows.

\section{Attention Comparison}
\label{sec:supp_attention_comparison}
\begin{figure*}
    \centering
    \includegraphics[width=1.0\linewidth]{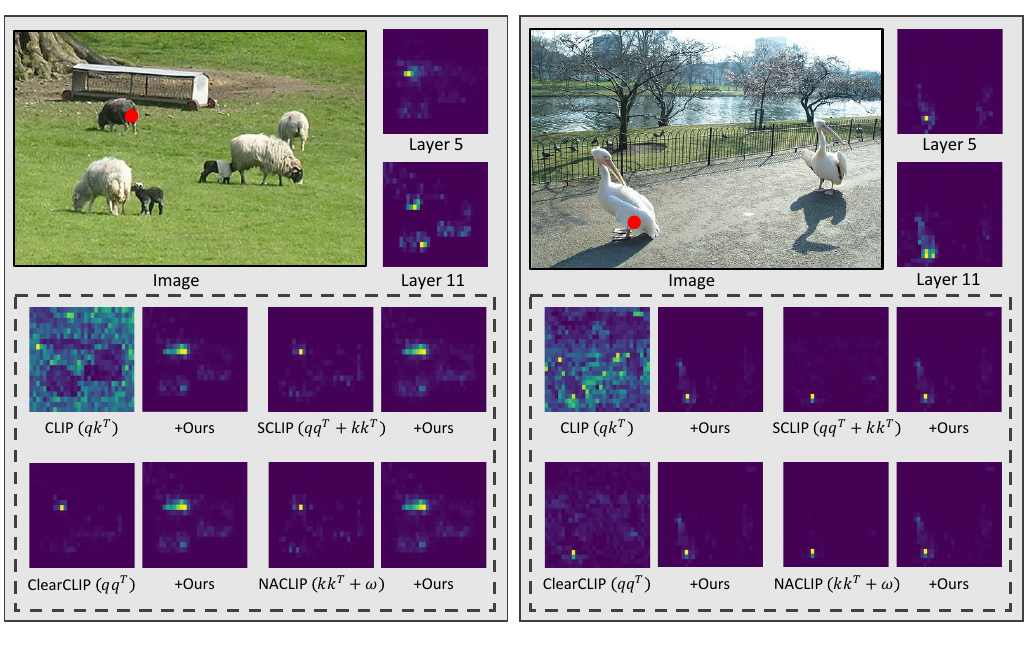}
    \vspace{-3mm}
    \caption{Additional comparison of attention maps across CLIP~\cite{radford2021learning}, SCLP~\cite{wang2025sclip}, ClearCLIP~\cite{lan2024clearclip}, NACLIP~\cite{hajimiri2025naclip}, and ours. The attention maps of non-last layers show the localization properties and can heal the attention in the last layer. The red point serves as the source point from which the attention map is computed and visualized.}
    \label{fig:fig10_supp_ours_attn_motivation_more}
    \vspace{2mm}
\end{figure*}
To further illustrate the impact of ResCLIP on attention mechanisms beyond examples shown in Fig.~3 in main paper, we present additional attention visualizations in Fig.~\ref{fig:fig10_supp_ours_attn_motivation_more}. These visualizations demonstrate how our method enhances the attention maps across different training-free open-vocabulary semantic segmentation (OVSS) models so that our method could better aggregate information from previous layers. From Fig.~\ref{fig:fig10_supp_ours_attn_motivation_more}, we can observe that our ResCLIP could attend to regions sharing similar class-specific features while previous works usually exhibit spatial-invariant features or focus on the local patches.

In particular, after integrating our Residual Cross-correlation Self-attention (RCS) and Semantic Feedback Refinement (SFR) modules into existing works, the attention maps show two key improvements: 1) enhanced local patches awareness and 
2) strengthened global semantic correspondence. For example, in the left part of Fig.~\ref{fig:fig10_supp_ours_attn_motivation_more}, we observe that previous works fail to effectively capture features from other ``sheep'' instances while our method can not only capture information from semantically consistent objects but also maintain local consistency. Similar phenomena can be observed from the right example in Fig.~\ref{fig:fig10_supp_ours_attn_motivation_more}. Moreover, we can see that intermediate layers (\eg, layers 5 and 11) show decent class-specific feature correspondence ability, which motivates us to incorporate them to remold the attention in the last block of CLIP. 

\section{Ablation Studies on ViT Backbones}
\label{sec:supp_ablation_studies_on_backbones}
\begin{table*}[t]
    \centering
    \caption{Ablation studies of our proposed modules in ViT-B/16, ViT-B/32 and ViT-L/14 backbones. Our ResCLIP setting is marked in \colorbox{Gray!16}{gray}. The best result on each dataset is \textbf{bolded}. The $\text{Avg}_{w/o}$ means the average mIoU for datasets \textit{without} a background class, $\text{Avg}_{w/}$ means the average mIoU for datasets \textit{with} a background class, and Avg. means the average mIoU for all eight datasets. 
    }
    \tablestyle{3.6pt}{1.0}
    \begin{tabular}{c|cc|ccc|ccc|ccc}
    \toprule
        \multirow{2}{*}{Methods} & \multicolumn{2}{c|}{Module} & \multicolumn{3}{c|}{ViT-B/16} & \multicolumn{3}{c|}{ViT-B/32} & \multicolumn{3}{c}{ViT-L/14}\\ 
         &  \textit{RCS} & \textit{SFR} & $\text{Avg}_{w/o}$ & $\text{Avg}_{w/}$ & Avg. & $\text{Avg}_{w/o}$ & $\text{Avg}_{w/}$ & Avg. & $\text{Avg}_{w/o}$ & $\text{Avg}_{w/}$ & Avg. \\
         \hline
        SCLIP~\cite{wang2025sclip} & - & - & 37.1 & 40.0 & 38.2 & 32.1 & 36.2 & 33.6 & 23.6 & 30.5 & 26.2 \\
        \hline
        \multirow{3}{*}{\makecell[c]{+ResCLIP(Ours)}}
        &  \checkmark & & 38.8 & 42.4 & 40.2 & 34.6 & 36.9 & 35.4 & 36.6 & 36.9 & 36.7 \\
        &  & \checkmark & 37.9 & 42.0 & 39.4 & 32.2 & 36.4 & 33.8 & 28.9 & 30.5 & 29.5 \\
         & \cellcolor{Gray!16}\checkmark &
        \cellcolor{Gray!16}\checkmark & \cellcolor{Gray!16}\textbf{39.3} & \cellcolor{Gray!16}\textbf{42.7} & \cellcolor{Gray!16}\textbf{40.5}
        & \cellcolor{Gray!16}\textbf{34.8} & \cellcolor{Gray!16}\textbf{37.1} & \cellcolor{Gray!16}\textbf{35.7} 
        & \cellcolor{Gray!16}\textbf{36.7} & \cellcolor{Gray!16}\textbf{37.4} & \cellcolor{Gray!16}\textbf{37.0} \\
        
        \hline
        ClearCLIP~\cite{lan2024clearclip} & - & - & 37.5 & 39.1 & 38.1 & 34.8 & 35.6 & 35.1 & 34.5 & 35.5 & 34.9 \\
        \hline
        \multirow{3}{*}{\makecell[c]{+ResCLIP(Ours)}}
        &  \checkmark & & 39.7 & 41.6 & 40.4 & 35.3 & 35.7 & 35.4 & 38.3 & 36.7 & 37.7 \\
        &  & \checkmark & 39.4 & 41.7 & 40.2 & 35.1 & 35.8 & 35.3 & 37.0 & 36.2 & 36.7 \\
         & \cellcolor{Gray!16}\checkmark &
        \cellcolor{Gray!16}\checkmark & \cellcolor{Gray!16}\textbf{40.0} & \cellcolor{Gray!16}\textbf{42.0} & \cellcolor{Gray!16}\textbf{40.7} 
        & \cellcolor{Gray!16}\textbf{35.5} & \cellcolor{Gray!16}\textbf{35.9} & \cellcolor{Gray!16}\textbf{35.6} 
        & \cellcolor{Gray!16}\textbf{38.4} & \cellcolor{Gray!16}\textbf{37.2} & \cellcolor{Gray!16}\textbf{37.9} \\
        
        \hline
        NACLIP~\cite{hajimiri2025naclip} & - & - & 38.2 & 41.4 & 39.4 & 34.4 & 37.0 & 35.4 & 36.2 & 36.9 & 36.5 \\
        \hline
        \multirow{3}{*}{\makecell[c]{+ResCLIP(Ours)}}
        &  \checkmark & & 39.7 & 42.2 & 40.6 & 35.7 & 37.3 & 36.3 & 38.4 & 38.2 & 38.3 \\ %
        &  & \checkmark & 39.3 & 42.9 & 40.7 & 35.7 & 37.4 & 36.3 & 37.4 & 38.4 & 37.8 \\ %
         & \cellcolor{Gray!16}\checkmark &
        \cellcolor{Gray!16}\checkmark & \cellcolor{Gray!16}\textbf{40.3} & \cellcolor{Gray!16}\textbf{43.2} & \cellcolor{Gray!16}\textbf{41.4} 
        & \cellcolor{Gray!16}\textbf{36.2} & \cellcolor{Gray!16}\textbf{37.5} & \cellcolor{Gray!16}\textbf{36.7} 
        & \cellcolor{Gray!16}\textbf{39.1} & \cellcolor{Gray!16}\textbf{39.2} & \cellcolor{Gray!16}\textbf{39.1} \\
        \bottomrule
    \end{tabular}
    \label{tab:supp_1_main_backbones_avg}
\end{table*}
In the main manuscript, we demonstrate effectiveness of the proposed RCS and SFR modules on ViT-B/16 backbone. To further demonstrate their generalization on other ViT backbones, we conduct additional experiments of ablation studies across ViT-B/16, ViT-B/32, and ViT-L/14 backbones. Moreover, we also evaluate our ResCLIP method by integrating it with previous training-free counterparts, \ie, SCLIP~\cite{wang2025sclip}, ClearCLIP~\cite{lan2024clearclip}, and NACLIP~\cite{hajimiri2025naclip}. 

The experimental results are shown in Table~\ref{tab:supp_1_main_backbones_avg}. We can see that both RCS and SFR modules contribute substantially to performance improvements across multiple backbones and baselines, demonstrating the great generalization of our proposed modules. Specifically, taking NACLIP with ViT-B/16 as an example, Our RCS improves the average mIoU from 39.4\% to 40.6\%, while SFR increases it to 40.7\%. When combining both modules, the performance further improves to 41.4\%, suggesting complementary benefits from both components. Similar patterns are observed with other baseline methods. 

Notably, our method demonstrates robust performance across different backbone architectures. For instance, when applied to SCLIP with ViT-L/14, ResCLIP significantly improves the average performance from 26.2\% to 37.0\%, showing particular effectiveness on larger architectures. The improvement is consistent across datasets both with and without a background class. Specifically, ViT-B/16 achieves 43.2\% mIoU on datasets with a background class, showing a 1.8\% mIoU improvement over NACLIP baseline, and 40.3\% mIoU on datasets without a background class, with a 2.1\% mIoU improvement. These comprehensive results validate that our proposed modules effectively enhance dense prediction capability of CLIP across various architectures and dataset configurations, demonstrating the robustness and generalization ability of our approach.

\section{Extension on other CLIP-like Models}
\label{sec:supp_extension_on_other_VLMs}
\begin{table*}[!t]
\centering
\caption{
Quantitative comparison on datasets \textit{without} a background class based on OpenCLIP~\cite{cherti2023reproducible} with ViT-B/16 architecture. Our results are marked in \colorbox{Gray!16}{gray}. The best results on each dataset are \textbf{bolded}. Results show that our method is also effective on other VLMs.
}
\vspace{-3mm}
  \tabcolsep3pt
  \centering
  \resizebox{0.7\textwidth}{!}{%
  \begin{tabular}{lcccccc}
    \toprule
    Methods  & VOC20 & Context59 & Stuff & Cityscape & ADE20k & Avg. \\
    \hline
    OpenCLIP \cite{cherti2023reproducible}  & 47.2 & 9.0 & 5.0 & 5.1 & 2.9 & 13.84 \\
    \hline
    SCLIP \cite{wang2025sclip}  & 66.6 & 31.7 & 21.2 & 31.4 & 18.5 & 33.9 \\
    \rowcolor{Gray!16}
    +ResCLIP(ours) & \textbf{71.8} & \textbf{32.9} & \textbf{21.9} & \textbf{31.9} & \textbf{18.8} & \textbf{35.5}
    \hfill\textcolor{ForestGreen}{(+1.6)}\\
 
    \hline
    ClearCLIP \cite{lan2024clearclip} & 81.4 & 34.1 & 23.1 & 31.8 & 18.9 & 37.9\\
    \rowcolor{Gray!16}
    +ResCLIP(ours) & \textbf{83.3} & \textbf{34.3} & \textbf{23.1} & \textbf{32.3} & \textbf{19.1} & \textbf{38.4}
    \hfill\textcolor{ForestGreen}{(+0.5)}\\

    \hline
    NACLIP \cite{hajimiri2025naclip}  & 76.2 & 30.3 & 20.3 & 32.3 & 17.6 & 35.3\\
    \rowcolor{Gray!16}
    +ResCLIP(ours)  & \textbf{82.5} & \textbf{33.0} & \textbf{22.2} & \textbf{32.9} & \textbf{19.0} & \textbf{37.9}
    \hfill\textcolor{ForestGreen}{(+2.6)}\\

  \bottomrule
  \label{tab:supp_3_different_clip_type_wobac}
  \end{tabular} 
  }
  \vspace{-5mm}
\end{table*}

In the main paper, we evaluate our method by integrating it with existing approaches, which are typically improved versions based on the vanilla CLIP model. To further evaluate the effectiveness of our method on other CLIP-like models, we conduct additional experiments on the OpenCLIP~\cite{cherti2023reproducible}. For a fair comparison, we first reproduce the results of SCLIP~\cite{wang2025sclip}, ClearCLIP~\cite{lan2024clearclip}, and NACLIP~\cite{hajimiri2025naclip} on OpenCLIP~\cite{cherti2023reproducible}. Then, we implement the proposed method based on the OpenCLIP~\cite{cherti2023reproducible}. As shown in Table~\ref{tab:supp_3_different_clip_type_wobac}, we present the comprehensive results on datasets without a background class. We can observe that our method shows consistent improvements over different baseline approaches, demonstrating its effectiveness. 

Specifically, when integrating SCLIP~\cite{wang2025sclip} with our method, ResCLIP achieves significant gains across all datasets, improving the average performance by 1.6\% mIoU. The improvement is particularly pronounced on VOC20, where ResCLIP enhances the mIoU from 66.6\% to 71.8\%. Most notably, integrating ResCLIP with NACLIP~\cite{hajimiri2025naclip} yields substantial improvements across all datasets, with an impressive average gain of 2.6\% mIoU, including a remarkable 6.3\% improvement on VOC20 datasets from 76.2\% to 82.5\%. These consistent improvements across different CLIP models and datasets demonstrate the generalization of our approach. The results also validate that the observation of our proposed method is effective on other CLIP-like models.

\section{Inference Efficiency Analysis}
\label{sec:supp_infer_efficiency}
The additional inference time introduced by our method is limited, as all operations only adjust attention in the final layer. Specifically, RCS computes the average of intermediate-layer attentions already generated during inference, while SFR performs lightweight mask adjustments in the final layer of CLIP. Using a single RTX 3090 GPU with batch size 1, input resolution of 336×336, and fp16 half precision, our experimental evaluation shows negligible impact on inference speed across all models. As shown in Table~\ref{tab:supp_effciency}, our enhancements increase total FLOPs by less than 7\%. Moreover, RCS demonstrates negligible overhead, while implementation of SFR can be further optimized to improve efficiency.
\begin{table}[h]
\caption{The Speed and FLOPs comparison of different methods on VOC 20 using CLIP-ViT-B/16 backbone. All the experiments are conducted on a single RTX 3090 GPU. IPS: Image Per Second.}
\renewcommand{\arraystretch}{1.0}
\centering
\resizebox{0.48\textwidth}{!}{
\begin{tabular}{c|cc|ccc}
\toprule
Metrics & CLIP & NACLIP & +\textit{RCS} & +\textit{SFR} & +ResCLIP \\
\hline
Speed (IPS) $\uparrow$ & 32.8 & 32.3 & 30.5 & 29.3 & 28.9 \\
FLOPs (G)  $\downarrow$ & 41.7 & 41.8 & 42.0 & 44.6 & 44.8 \\
\bottomrule
\end{tabular}
}
\label{tab:supp_effciency}
\end{table}

\section{Additional Visualization Results}
\label{sec:supp_seg_visual}
\begin{figure*}
    \centering
    \includegraphics[trim={0 4mm 0 0},clip]{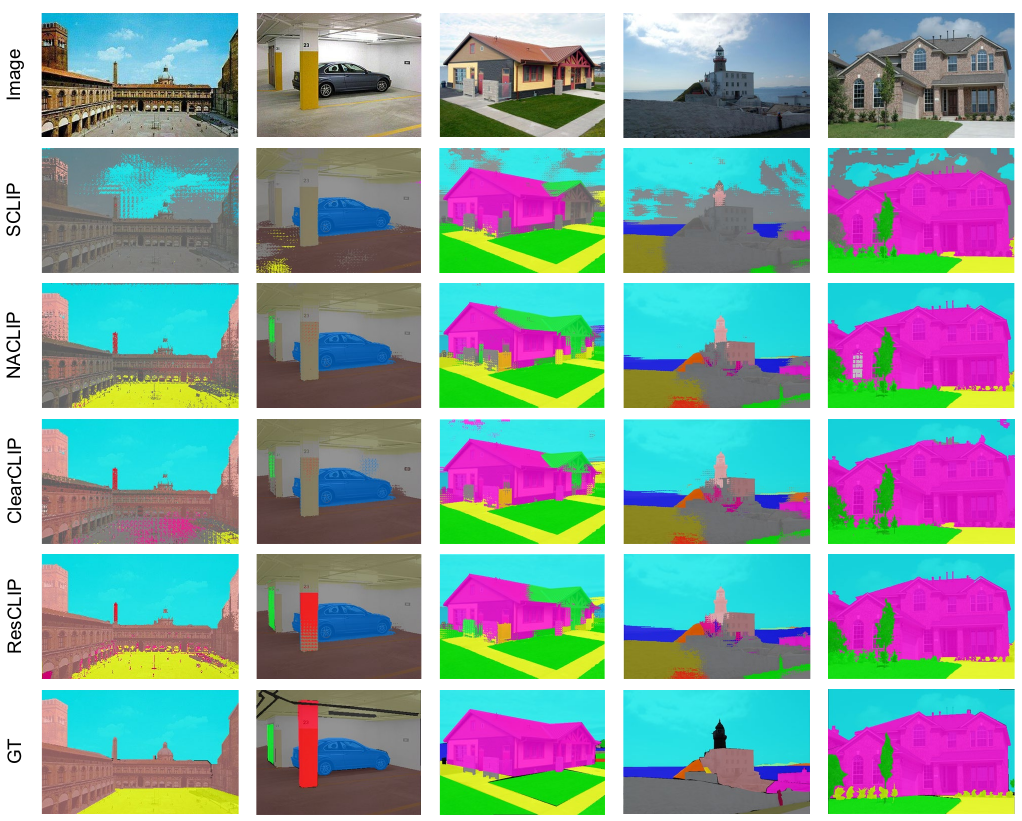}
    \vspace{-1mm}
    \caption{Additional qualitative visualization results among different CLIP-based training-free segmentation methods on ADE20K~\cite{zhou2019semantic} dataset.
    }
    \label{fig:fig7_experiment_visual_supp_ade20k}
    \vspace{-1mm}
\end{figure*}

\begin{figure*}
    \centering
    \includegraphics[trim={0 4mm 0 0},clip]{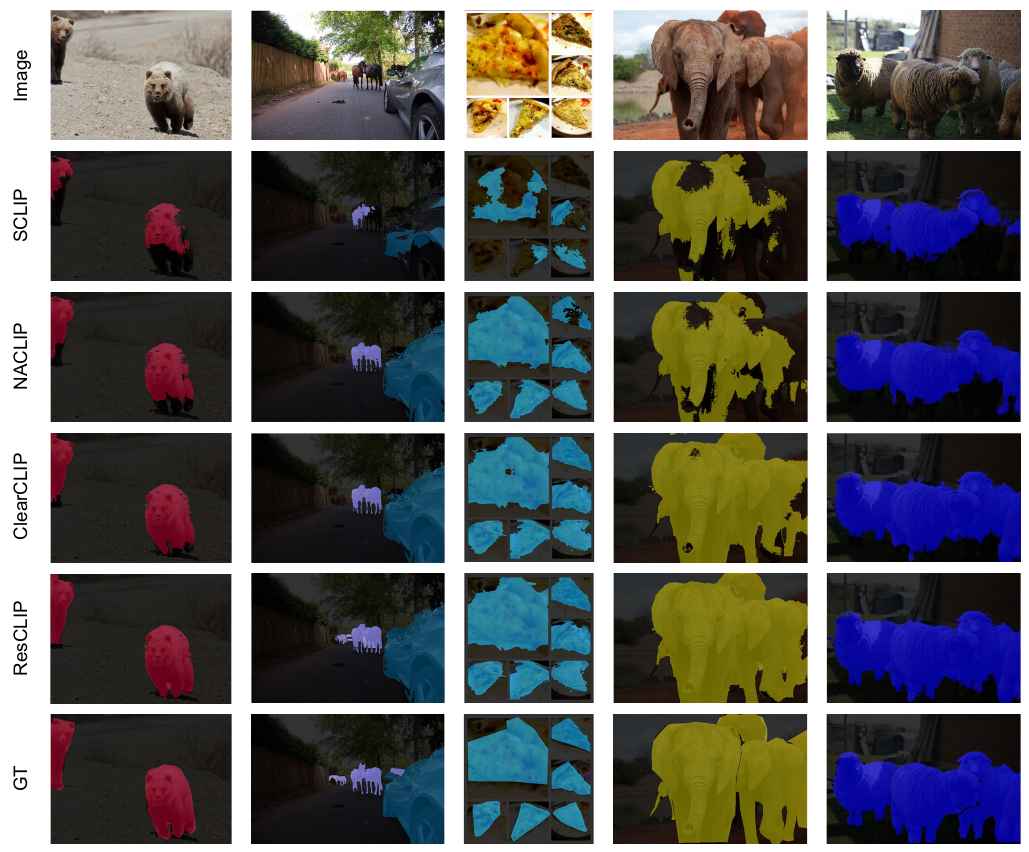}
    \vspace{-5mm}
    \caption{Additional qualitative visualization results among different CLIP-based training-free segmentation methods on COCO Object~\cite{caesar2018coco} dataset.
    }
    \label{fig:fig8_experiment_visual_supp_coco_obj}
    \vspace{-1mm}
\end{figure*}

\begin{figure*}
    \centering
    \includegraphics[trim={0 4mm 0 0},clip]{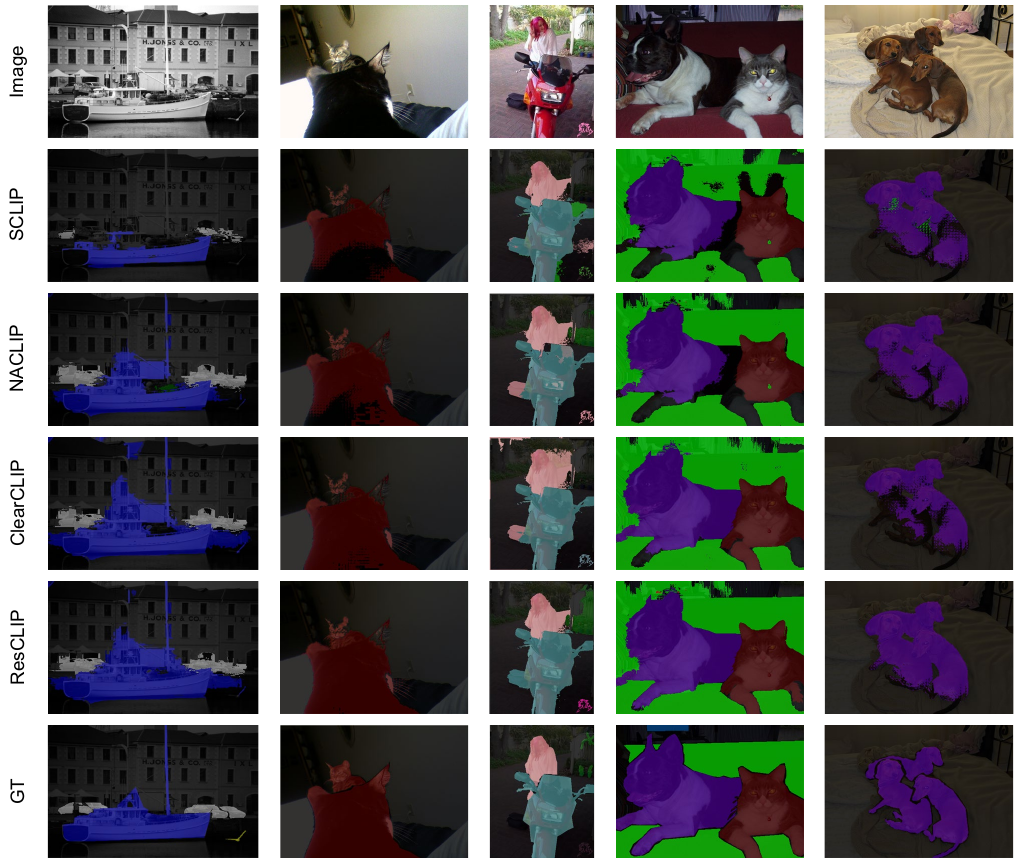}
    \vspace{-1mm}
    \caption{Additional qualitative visualization results among different CLIP-based training-free segmentation methods on PASCAL VOC~\cite{everingham2015pascal} dataset.
    }
    \label{fig:fig9_experiment_visual_supp_voc}
    \vspace{-1mm}
\end{figure*}

We present additional qualitative comparisons across ADE20K~\cite{zhou2019semantic}, COCO Object~\cite{caesar2018coco}, and PASCAL VOC~\cite{everingham2015pascal} datasets in Fig.~\ref{fig:fig7_experiment_visual_supp_ade20k}, Fig.~\ref{fig:fig8_experiment_visual_supp_coco_obj}, and Fig.~\ref{fig:fig9_experiment_visual_supp_voc} to further demonstrate the effectiveness of our ResCLIP, respectively. Compared to existing methods, our approach usually presents better quality in terms of the semantic segmentation masks. From these qualitative results, we can have the following observations: 1) 
Our method generates significantly cleaner segmentation masks with reduced noise artifacts. This improvement is particularly evident in complex scenes from ADE20K, where ResCLIP maintains coherent building segmentation without the internal hollows or fragmentations commonly seen in other baselines (\ie, the 1-st \textit{col.} in Fig.~\ref{fig:fig7_experiment_visual_supp_ade20k}). The enhanced segmentation quality extends to diverse scenarios, such as the precise delineation of vehicles in parking lots and the clear separation of multiple instances in crowded scenes (\ie, the 2-nd and 4-th \textit{col.} in Fig.~\ref{fig:fig7_experiment_visual_supp_ade20k}). 2) ResCLIP presents superior performance in handling multiple object instances, demonstrating its enhanced spatial-semantic understanding. For example, in the COCO Object dataset (see Fig.~\ref{fig:fig8_experiment_visual_supp_coco_obj}), our method accurately segments groups of animals while maintaining clear boundaries between individuals(\ie, the 4-th and 5-th \textit{col.} in Fig.~\ref{fig:fig8_experiment_visual_supp_coco_obj}). This capability stems from the improved attention mechanism of our ResCLIP, which better captures both global spatial relationships and local feature consistency. 3) Our method handles varying scales and perspectives better. As shown in Fig.~\ref{fig:fig9_experiment_visual_supp_voc}, our method produces consistent segmentation quality across both indoor and outdoor scenes. These qualitative results validate the effectiveness of our proposed RCS and SFR modules in enhancing dense prediction capabilities of CLIP. 

\end{document}